\pdfoutput=1
\documentclass{article}

\usepackage{microtype}
\usepackage{graphicx}
\usepackage{subfigure}
\usepackage{booktabs}
\usepackage{array}
\usepackage{amsmath,amssymb}
\usepackage{multirow}
\usepackage{xcolor}
\usepackage{tikz}
\usetikzlibrary{positioning,arrows.meta,fit,calc}
\usepackage{hyperref}

\usepackage{algorithm}
\usepackage{algorithmic}

\usepackage[accepted]{mlsys2025}
\makeatletter
\renewcommand{\Notice@String}{Preprint. Copyright 2026 by the author(s).}
\makeatother

\newcommand{\sysname}{\textsc{EviGate}}

\mlsystitlerunning{Claim Laundering: When Agent Decision Models Trust Records Over Reality}

\hypersetup{pdftitle={Claim Laundering: When Agent Decision Models Trust Records Over Reality},
  pdfauthor={Tina Khezresmaeilzadeh, Seyedarmin Azizi, Erfan Baghaei Potraghloo, Mehdi Kamal, Massoud Pedram}}

\begin{document}

\twocolumn[
\mlsystitle{Claim Laundering: When Agent Decision Models Trust Records Over Reality}
\chead{\small\bf Claim Laundering: When Agent Decision Models Trust Records Over Reality}  

\mlsyssetsymbol{equal}{*}
\begin{mlsysauthorlist}
\mlsysauthor{Tina Khezresmaeilzadeh}{usc}
\mlsysauthor{Seyedarmin Azizi}{usc}
\mlsysauthor{Erfan Baghaei Potraghloo}{usc}
\mlsysauthor{Mehdi Kamal}{usc}
\mlsysauthor{Massoud Pedram}{usc}
\end{mlsysauthorlist}
\mlsysaffiliation{usc}{University of Southern California, Los Angeles, CA, USA}
\mlsyscorrespondingauthor{Tina Khezresmaeilzadeh}{khezresm@usc.edu}
\mlsyskeywords{LLM Agents, Decision Models, Silent Failures, Interpretability, Probing, Activation Patching}

\vskip 0.3in

\begin{abstract}
Agent decision models select actions in a single forward pass from inputs that combine environmental observations with the agent's own record of previous actions. When an action fails silently, these sources can conflict: the record reports success, but the environment shows that the step did not take effect. We ask which source determines the next decision. Across Android screen traces, ALFWorld, and WebShop, we construct paired states that independently vary the observation, the logged action, and an explicit success claim. In all 12 configurations of four decision models and three domains, records have a larger average effect on commitment decisions than observations; after claimed-but-failed steps, the models still stop, advance, or purchase in 53--100\% of cases. Yet the relevant evidence is often represented internally: linear probes on intermediate hidden states distinguish successful from failed steps on held-out apps, tasks, and goals (AUROC 0.77--1.00), and activation patching reveals increasing influence from records in later layers. We use this signal in \sysname{}, an intermediate-layer veto that requires no additional forward pass and adds 0.04--2.0\% serving latency. Under injected silent failures, \sysname{} improves task success by 18.7 points in live ALFWorld agents, 5.0 in WebShop, and 5.1 in a phone simulator, and in WebShop and the simulator it also outperforms random vetoes at the same rate. Its effectiveness depends on representative calibration states: gates calibrated on offline counterfactual states do not transfer reliably to live agent trajectories.
\end{abstract}
]

\printAffiliationsAndNotice{}

\section{Introduction}

Agents spend most of their inference on small, repeated decisions: stop or continue, buy or keep looking, move on or retry.
A new class of \emph{System-One decision models} is built for exactly this
workload~\cite{typesafe2026jev,palmer2026kev,marosi2026decider,convai2026laya}. Given a state and a typed question, such a
model returns a probability for every predefined option in a single forward pass, at a fraction of the latency and cost of
LLM generation. Jev is already served through OpenRouter~\cite{openrouterjev} and Cloudflare Workers AI and integrated with
LangChain and Pydantic AI.

\paragraph{The problem: records can disagree with the world.} A decision model does not watch the environment. It reads a
state that the agent assembles: the goal, a record of what the agent did (LLM-written step summaries in T3A, ReAct thoughts
in ALFWorld, logged clicks in WebShop) and the current observation. The record and the world can diverge. Actions fail
silently: a tap misses, a click does not register, a command returns ``Nothing happens.''. And records are written from the
action's \emph{intent}, often before its effect can be checked. A 9B summarizer given T3A's own prompt, which asks it to ``be
critical'', still reports success after 39\% of actions that had no effect (\S\ref{sec:measure}). When the record and the
observation disagree, the observation is right. A reliable decision layer must notice the mismatch before it takes an
irreversible step. One that follows the record instead turns the agent's claim into the decision; we call this
\emph{claim laundering} (Figure~\ref{fig:teaser}).

This matters in practice. Replacing an LLM action selector with Jev cut the cost of every decision but lowered end-to-end
success on AndroidWorld by 13--18 points, with premature completion among the leading causes~\cite{ling2026faster}. Across
LLM agents, 45--76\% of failures on $\tau^2$-bench and AppWorld are \emph{false successes}, where the agent reports
completion the environment does not show~\cite{falsesuccess2026}. These studies describe the symptom. We ask where it comes
from in the component that makes the decision, and how to detect it.

\paragraph{What we do.} We answer four questions, each in its own section.
\begin{enumerate}
\item \emph{Do decision models detect record--state mismatches?} (\S\ref{sec:measure}) We build paired states that change
only the record or only the observation, in three domains, and test seven open decision models, hosted Jev and their base
LLMs. They do not: the record moves the decision more than the observation for all four models tested in every domain,
and for six of the seven open models on phones.
\item \emph{Does the model know better than it says?} (\S\ref{sec:latent}) Yes. A linear probe on the hidden state detects
whether the step worked, often where the model's own output is near chance.
\item \emph{Why does the output ignore it?} (\S\ref{sec:why}) Activation patching shows the observation reaching the
decision position by mid-depth, while record-dependent influence becomes dominant in the last layers.
\item \emph{Can we use this to detect mismatches?} (\S\ref{sec:detector}--\S\ref{sec:loops}) \sysname{} reads the middle
layer and vetoes unsupported stop, proceed or buy decisions. We test it offline and in three live agents, against random
vetoing at the same rate. We then ask whether simply fine-tuning the model would
be enough (\S\ref{sec:finetune}).
\end{enumerate}

\paragraph{Contributions.}
\begin{itemize}
\item \emph{A failure and how to measure it.} We define record--state mismatch for agent decision layers and measure it with
counterfactual pairs in three domains. After a failed step whose record reports success, the models still stop, proceed or
buy in 53--100\% of cases. The bias is already present in the base LLM, and decision tuning does not remove it.
\item \emph{Where it happens.} The mismatch is encoded in the hidden state (probe AUROC 0.77--1.00 on held-out groups) and
outweighed by record-dependent influence in the last layers.
\item \emph{A detector that works in live agents.} \sysname{} costs one dot product and raises task success in all three
domains, provided it is calibrated on states from the deployed agent itself. Fine-tuning removes the bias too, but only in the domains
it is trained on.
\end{itemize}

\section{Background}
\label{sec:background}
\begin{figure}[t]
\centering\small
\fbox{\parbox{0.95\columnwidth}{\ttfamily\scriptsize
Goal: Turn on Wi-Fi\\
Action history:\\
\ \ Step 3: tapped 'Wi-Fi toggle'\\
\ \ \textcolor{red!70!black}{Agent note: successfully completed: Turn on Wi-Fi.}\\
Current screen (Settings > Network):\\
\ \ \textcolor{blue!60!black}{- Wi-Fi toggle: Off}\\
\ \ - Text: Search settings \ldots
}}\\[3pt]
{\footnotesize
\begin{tabular}{@{}l@{\hskip 4pt}l@{}}
Decision model output: & $p(\texttt{stop})=0.65$ $\Rightarrow$ \textbf{declares done} \\
\sysname{} (layer 20): & $p(\text{step worked})=0.00$ $\Rightarrow$ \textbf{veto} \\
\end{tabular}}
\caption{A record--state mismatch (real instance from our diagnostic set). The agent's record (red) says the task is done;
the screen (blue) shows it is not. The decision model follows the record. A probe on its own hidden state, trained on
\emph{other} goals, reads the screen correctly.}
\label{fig:teaser}
\end{figure}
\paragraph{Decision models.} A request contains a \emph{state} (text or JSON) and typed \emph{questions}, for example a
\texttt{choice} among named options. The model reads state and questions in one sequence and returns a probability for every
option in one forward pass. The open models differ in how they read the answer. \textbf{decider}~\cite{marosi2026decider}
fine-tunes Qwen3.5 base models~\cite{qwen35} (Gemma-4 for its 12B model) and reads option-letter logits at an answer
position. \textbf{Kev}~\cite{palmer2026kev} keeps a Qwen3.5 backbone frozen, adds a rank-16 LoRA and scores options with a
pointer head. \textbf{Laya}~\cite{convai2026laya} is a ModernBERT encoder~\cite{warner2025modernbert}. The hosted Jev model is
closed-weight; all three open families serve the same API.

\paragraph{Decision models in agents.} \citet{ling2026faster} integrate Jev into two agents, and we reuse both request
formats verbatim. In the Android agent T3A~\cite{rawles2024androidworld}, the state is T3A's full prompt (goal, LLM-written
step summaries, list of screen elements) and one question chooses among twelve operations, including \texttt{complete}. In
ReAct~\cite{yao2023react} on ALFWorld~\cite{shridhar2021alfworld}, the state is the ReAct transcript and the question chooses
the next command. We apply the same pattern to WebShop~\cite{yao2022webshop}.

\section{Measuring Record--State Mismatch}
\label{sec:setup}
\paragraph{Problem formulation.}
A decision model receives a goal, an agent-generated action record, the latest environmental observation, and a typed question. We write this input as $x=(g,r,o,q)$ and denote by $p_\theta(a^\star\mid x)$ its probability for the decision $a^\star$ that commits to the preceding step having succeeded. The observation $o$ describes the current environment; the record $r$ describes what the agent believes or says it did. These are not interchangeable. An attempted action may be logged even if it has no effect, and an agent may explicitly claim success despite an unchanged environment. We call a case a \emph{record--state mismatch} when the record suggests an achieved outcome that the observation does not support. An explicit success claim creates a direct contradiction; a truthful log of an attempted action may instead become a misleading cue when the model treats the attempt as evidence of success.

We distinguish two parts of the record: a \emph{logged action}, which states that an operation was attempted, and an \emph{explicit success claim}, which states that its intended outcome occurred. We say a step \emph{took effect} when the environment reached the intended post-action state. The decision $a^\star$ is \texttt{complete} on phones, advancing past a required subgoal in ALFWorld, and \texttt{buy now} in WebShop. These are consequential commitments, although advancing in ALFWorld is not literally an episode-terminal action.

\paragraph{Counterfactual pairs.}
To isolate the influence of each input, we construct contrast sets~\cite{gardner2020contrast,kaushik2020cad} around decision points at which a step should complete a task or a required subgoal. Each set contains five variants of the same example, with the goal, question, and unrelated context held fixed:
\begin{itemize}
\item \textsc{Done}: the action is logged and the observation shows its intended effect, without an explicit success claim.
\item \textsc{Done+claim}: the same successful transition, with an explicit success claim.
\item \textsc{Honest}: the action has not been attempted, is not logged, and the observation shows the preceding state.
\item \textsc{Logged}: the action is logged but has no effect; the observation still shows the preceding state.
\item \textsc{Logged+claim}: the same failed transition, with an explicit but incorrect success claim.
\end{itemize}
The variants support three controlled comparisons. Writing $L(v)=\log p_\theta(a^\star\mid x_v)$, we measure
\begin{align}
\Delta_{\mathrm{obs}} &= L(\textsc{Done})-L(\textsc{Logged}),\\
\Delta_{\mathrm{log}} &= L(\textsc{Logged})-L(\textsc{Honest}),\\
\Delta_{\mathrm{claim}} &= L(\textsc{Logged+claim})-L(\textsc{Logged}).
\end{align}
The first changes the observed outcome while keeping the action log fixed; the second adds an attempted action without changing the observed outcome; the third adds a success assertion to a failed attempt. Thus a positive effect increases the model's support for committing to completion. Effects are differences in \emph{log probabilities}: for example, $+0.7$ corresponds to approximately twice the probability, not a 70-percentage-point increase. We also compare \textsc{Done+claim} with \textsc{Logged+claim} when testing whether a detector distinguishes success from failure despite an identical success claim.

\begin{table}[t]
\caption{The three domains: real Android screens from AndroidControl~\cite{li2024androidcontrol} (245 apps), unseen
ALFWorld tasks, and WebShop goals. Request formats follow \citet{ling2026faster}.}
\label{tab:domains}
\centering\footnotesize
\setlength{\tabcolsep}{2.5pt}
\begin{tabular}{@{}lllr@{}}
\toprule
Domain & Terminal decision & Silent failure & States \\
\midrule
Phones & \texttt{complete} & screen unchanged & 3{,}200 \\
ALFWorld & proceed & ``Nothing happens.'' & 1{,}475 \\
WebShop & \texttt{buy now} & options not selected & 1{,}500 \\
\bottomrule
\end{tabular}
\end{table}

\paragraph{Three domains.}
Table~\ref{tab:domains} summarizes 6{,}175 states from 1{,}235 decision points. \emph{Phones}: we use screen-element descriptions and action histories from real AndroidControl trajectories~\cite{li2024androidcontrol}, spanning 245 apps. The screen after a task-completing action supplies the successful observation (we assume, as the dataset intends, that the demonstrated final action completes the task); the preceding screen supplies the unchanged observation in the counterfactual failure. The record consists of step summaries, with an optional completion claim. The decision model reads the textual representation of a real screen, not raw image pixels. \emph{ALFWorld}: at up to three state-changing steps along expert trajectories, we use the environment's ``Nothing happens.'' feedback for unsuccessful attempts and vary whether the ReAct transcript contains the attempted step and a success thought. \emph{WebShop}: the agent reaches the target product through environment interactions and attempts to select the required options; for failed variants, those clicks remain recorded although the current page shows the options unselected. The construction and domain-specific prompts are detailed in Appendix~\ref{app:details}.

These paired failures are controlled counterfactuals, not estimates of how often silent failures occur during unmodified deployment. This distinction allows us to measure \emph{which input drives a decision} without confounding the result with the natural frequency of action failures.

\paragraph{Models and evaluation.}
Four models are evaluated in all three domains: decider-4b, Kev-4B, decider-12b, and hosted Jev (\texttt{jev-1.13-20260917}). On phones, we additionally evaluate decider-2b, Kev-0.8B, Kev-9B, and Laya, as well as the base and chat Qwen3.5-4B models underlying the 4B decision models. We preserve the agent request formats of~\citet{ling2026faster}; only the targeted input component changes within each comparison. We report the proportion of \textsc{Logged+claim} cases in which the model chooses $a^\star$, alongside mean paired log-probability effects. Confidence intervals are obtained by bootstrapping over episodes or evaluation groups rather than treating variants of a single decision point as independent.

\paragraph{Hidden-state probes.}
For each open model, we extract the hidden state at the answer position every two layers and fit an $\ell_2$-regularized logistic probe to predict whether the step took effect. All variants of a decision point remain in the same split; evaluation holds out entire apps, tasks, or goals using GroupKFold. Layer selection takes place by cross-validation within training groups, not on test groups. We compare the probe's AUROC with AUROC from the model's own output score, both on all states and on the subset whose records claim success. The latter comparison tests whether the observation remains decodable when the claim itself provides no information about the label.

\section{Decision Models Favor Records Over Observations}
\label{sec:measure}
\begin{table}[t]
\caption{Which input moves the decision. \emph{Wrong}: how often the model still stops, proceeds or buys after a failed step
whose record claims success (\textsc{Logged+claim}). \emph{Obs.}, \emph{Logged}, \emph{Claim}: how much the observation,
the logged step and the explicit claim raise $\log p$ of that option. \textbf{Bold}: the largest of the three. In every row
the bold record effect exceeds the observation effect with a paired 95\% CI above zero (Appendix Table~\ref{tab:effectsci}).}
\label{tab:effects3}
\centering\small
\setlength{\tabcolsep}{4pt}
\begin{tabular}{@{}lcccc@{}}
\toprule
Model & Wrong & Obs. & Logged & Claim \\
\midrule
\multicolumn{5}{@{}l}{\emph{Phones}} \\
decider-4b & 95\% & $+.05$ & $\mathbf{+1.10}$ & $+.80$ \\
Kev-4B & 96\% & $+.09$ & $+.20$ & $\mathbf{+.32}$ \\
decider-12b & 84\% & $+.21$ & $\mathbf{+1.10}$ & $+.56$ \\
hosted Jev & 88\% & $+.61$ & $\mathbf{+2.31}$ & $+.95$ \\
\midrule
\multicolumn{5}{@{}l}{\emph{Embodied (ALFWorld)}} \\
decider-4b & 90\% & $+1.19$ & $\mathbf{+2.60}$ & $+1.28$ \\
Kev-4B & 80\% & $+.28$ & $+.25$ & $\mathbf{+.32}$ \\
decider-12b & 97\% & $+.40$ & $\mathbf{+1.20}$ & $+.90$ \\
hosted Jev & 95\% & $+.98$ & $\mathbf{+4.41}$ & $+1.16$ \\
\midrule
\multicolumn{5}{@{}l}{\emph{Web (WebShop)}} \\
decider-4b & 100\% & $+.02$ & $\mathbf{+3.71}$ & $+.07$ \\
Kev-4B & 53\% & $+.11$ & $\mathbf{+.31}$ & $+.23$ \\
decider-12b & 99\% & $-.09$ & $\mathbf{+1.20}$ & $+.98$ \\
hosted Jev & 99\% & $+.20$ & $\mathbf{+2.75}$ & $+.57$ \\
\bottomrule
\end{tabular}
\end{table}

\paragraph{Records dominate the controlled comparisons.}
Table~\ref{tab:effects3} reports the four models evaluated in every domain. In all 12 model--domain configurations, the largest \emph{mean} increase in the probability of the commitment decision comes from a record component (the logged action or explicit claim), not the environmental observation. This is a comparison of causal input edits within matched examples; it does not imply that every individual decision follows the record. The comparison is statistically reliable: in every configuration the larger record effect exceeds the observation effect with a paired 95\% bootstrap interval above zero, including the closest case, Kev-4B in ALFWorld (claim minus observation $+0.04$ [$+0.01$, $+0.07$]). When the false claim is present, changing the observation from failed to successful moves the decision by at most $+0.24$, and by $+0.035$ or less for decider-4b and decider-12b on phones and WebShop (Appendix Table~\ref{tab:effectsci}).

On \emph{phones}, an attempted action recorded without its intended effect increases the log probability of \texttt{complete} by $+0.20$ to $+2.31$, whereas showing that the screen actually reached the target state contributes only $+0.05$ to $+0.61$. With both the unsuccessful action and a false success claim in the record, these models stop in 84--96\% of cases. In \emph{ALFWorld}, the observation explicitly says ``Nothing happens.'' Nevertheless, the models advance past the unfinished subgoal in 80--97\% of \textsc{Logged+claim} cases. In \emph{WebShop}, the current page shows the required options unselected, but the logged clicks raise the log probability of purchasing by $+0.31$ to $+3.71$, compared with an observation effect between $-0.09$ and $+0.20$. The resulting wrong-purchase rate is 53--100\% across models. Hosted Jev exhibits the same qualitative pattern as the open models, so it is not confined to one open-model architecture or answer head.

\paragraph{The effect persists with LLM-written records.}
The preceding explicit claims are controlled templates. To test whether a realistic summarization component produces misleading records, we run Qwen3.5-9B with T3A's own step-summary prompt, providing the before- and after-step screen descriptions in the phone failure condition. Despite access to the unchanged observation, 39\% of these generated summaries (78 of 200) report that the action succeeded (Appendix~\ref{app:details}). Conditional on such a false summary, decider-4b, Kev-4B, and hosted Jev choose \texttt{complete} in 77\%, 94\%, and 80\% of cases, respectively; with an honest summary, the corresponding rates are 26\%, 48\%, and 27\%. Thus an incorrect LLM-written record increases premature completion by 46--53 percentage points in these paired evaluations. The 39\% figure measures summarizer behavior on our failed-step test cases, not the prevalence of failure in deployed Android agents. In the live ALFWorld and WebShop experiments (\S\ref{sec:loops}), the records are instead generated by the acting agents themselves.

\paragraph{Neither scale nor decision tuning consistently removes the bias.}
The additional phone models exhibit the same general vulnerability (Appendix Table~\ref{tab:effects}). For models up to 4B, including the encoder Laya, the explicit claim has a greater mean effect than the observed screen transition. Kev-9B is a partial exception: the screen contributes $+0.34$ compared with $+0.27$ for the log and $+0.24$ for the claim, yet the model still completes after 65\% of claimed failures. We also compare the decision-tuned models with their Qwen3.5-4B backbone (Appendix Table~\ref{tab:backbone}). The base model already responds more strongly to the claim ($+0.51$) than to the observed outcome ($+0.02$; not significant). In these comparisons, decision tuning does not eliminate that behavior: Kev's logged-action effect changes from $+0.15$ to $+0.21$, while decider's increases to $+1.12$. The results are consistent with a preference for action records that predates decision-specific tuning, although they do not by themselves identify its training-time cause.

The behavioral evidence establishes a failure of \emph{evidence use}, rather than simply a lack of information: models are given an observation that contradicts the record, but their selected commitment is disproportionately driven by what the record says. We next test whether the contradiction is recoverable from the representations that produce those decisions.

\section{Hidden States Retain Evidence of Failure}
\label{sec:latent}
\begin{table}[t]
\caption{Detecting whether the step worked (AUROC on held-out apps, tasks or goals; 0.5 is chance). \emph{Head}: the
model's own output. \emph{Probe}: a linear probe on its hidden state. \emph{All}: all test states. \emph{Claimed}: only
states whose record claims success, so that only the observation differs. Jev is closed-weight and cannot be probed.}
\label{tab:probe3}
\centering\small
\setlength{\tabcolsep}{5pt}
\begin{tabular}{@{}lcc@{\hskip 10pt}cc@{}}
\toprule
& \multicolumn{2}{c}{All} & \multicolumn{2}{c}{Claimed} \\
\cmidrule(lr){2-3}\cmidrule(l){4-5}
Model & Head & Probe & Head & Probe \\
\midrule
\multicolumn{5}{@{}l}{\emph{Phones}} \\
decider-4b & .63 & $\mathbf{.77}$ & .51 & $\mathbf{.74}$ \\
Kev-4B & .65 & $\mathbf{.80}$ & .55 & $\mathbf{.79}$ \\
decider-12b & .69 & $\mathbf{.80}$ & .61 & $\mathbf{.77}$ \\
hosted Jev & .75 & -- & .68 & -- \\
\midrule
\multicolumn{5}{@{}l}{\emph{Embodied (ALFWorld)}} \\
decider-4b & .92 & $\mathbf{1.00}$ & .96 & $\mathbf{1.00}$ \\
Kev-4B & .78 & $\mathbf{1.00}$ & .80 & $\mathbf{1.00}$ \\
decider-12b & .72 & $\mathbf{1.00}$ & .72 & $\mathbf{1.00}$ \\
hosted Jev & .83 & -- & .89 & -- \\
\midrule
\multicolumn{5}{@{}l}{\emph{Web (WebShop)}} \\
decider-4b & .60 & $\mathbf{1.00}$ & .54 & $\mathbf{1.00}$ \\
Kev-4B & .72 & $\mathbf{1.00}$ & .65 & $\mathbf{1.00}$ \\
decider-12b & .55 & $\mathbf{1.00}$ & .53 & $\mathbf{1.00}$ \\
hosted Jev & .54 & -- & .52 & -- \\
\bottomrule
\end{tabular}
\end{table}

\paragraph{The hidden representation is more informative than the final score.}
Table~\ref{tab:probe3} compares two ways to predict whether the last step took effect: the decision model's own commitment score (\emph{Head}) and a linear readout of its internal state (\emph{Probe}). Both are evaluated on held-out apps, tasks, or goals. The most diagnostic setting is \emph{Claimed}: successful and unsuccessful states contain the same success assertion, so a classifier cannot obtain the label simply by detecting optimistic record language. It must exploit information associated with the environment's actual state.

On phones, where the action outcome must be inferred from screen-element descriptions, the open-model heads reach only 0.51--0.61 AUROC on claimed states. Linear probes reach 0.74--0.79, improving discrimination by 0.16--0.24 AUROC points. On WebShop, the heads achieve 0.53--0.65 while probes reach 1.00. In ALFWorld, the heads perform better (0.72--0.96), but the probes again reach 1.00. The hosted Jev outputs exhibit the same limitations in some settings (0.68 AUROC on claimed phone states and 0.52 on claimed WebShop states), although Jev's hidden states are unavailable for probing. Across all states, open-model probe AUROC ranges from 0.77 to 1.00.

\paragraph{Separating useful representations from obvious textual cues.}
Perfect probing on ALFWorld and WebShop should not be mistaken for evidence of sophisticated hidden reasoning. Their failures are accompanied by distinctive, consistent textual traces; a classifier trained directly on the input text also reaches 1.00 AUROC. The important observation in those domains is narrower: the outcome is readily available, yet the decision head often fails to use it. Phones provide the more demanding test. Failure is implicit in longer screen descriptions, and the text-only classifier reaches 0.59 AUROC, compared with 0.77--0.80 for the hidden-state probes on all states. Thus the intermediate representations expose a more useful signal than this text baseline or the model's own commitment score, including on held-out apps.

\paragraph{Where the signal appears and how well it transfers.}
In the 32-layer 4B models, phone outcome information becomes linearly readable around layer 16, with the strongest probes at roughly 55--75\% of network depth (Appendix Figure~\ref{fig:layers}). A probe trained on only ten labelled phone episodes already outperforms the model's final score (Appendix Figure~\ref{fig:lc}). But representation quality does not make calibration domain-independent: training on synthetic template states and evaluating on real phone-screen descriptions yields only 0.63 AUROC. This shortfall motivates fitting the eventual detector on labelled states representative of the target agent, rather than relying on mismatched synthetic examples.

These findings establish a \emph{decodability gap}: relevant evidence about action outcomes is present in an intermediate representation but is not reliably reflected in the model's output probability. Linear decodability alone does not show how the model uses that information or why its output favors the record. We therefore turn to activation patching (\S\ref{sec:why}) to locate the layerwise contributions to the decision, before using an intermediate-layer readout as a gate (\S\ref{sec:detector}).

\section{Why the Output Ignores It}
\label{sec:why}
\begin{figure*}[t]
\centering
\includegraphics[width=0.82\textwidth]{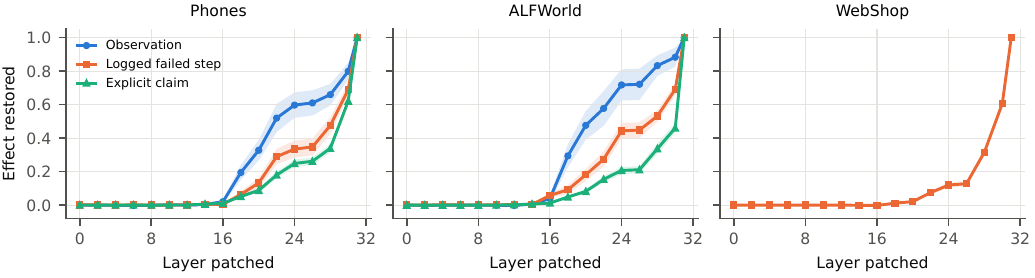}
\caption{Activation patching on decider-4b: the share of each input's effect on $\log p$ of the terminal option that is
restored by copying the decision-position state from the paired state at a single layer (mean $\pm$95\% CI). The
observation is present by mid-depth; the logged step and the claim arrive in the last layers. We keep pairs whose $\log p$
differs by more than 0.2. In WebShop the page barely moves the decision ($+0.02$, Table~\ref{tab:effects3}), so there is no
observation effect to locate and only the logged-click curve ($n=100$) is shown.}
\label{fig:mech3}
\end{figure*}

To find where the record takes over, we use activation patching. We run a state, replace the hidden state at the decision
position in one layer with the one from its paired state (e.g.\ \textsc{Logged} in place of \textsc{Logged+claim}), and
measure how much of the difference in $\log p$ this restores. Patching costs 19 forward passes per pair, so we run
it on decider-4b over 100 decision points per domain (Figure~\ref{fig:mech3}) and repeat it on decider-12b
(Appendix~\ref{app:patch12b}). The pattern is the same in every domain:
\begin{itemize}
\item \emph{The observation arrives at mid-depth.} At layer 24 of 32, patching restores 60\% (phones) and 72\% (embodied) of
the observation's effect.
\item \emph{The record arrives late.} At layer 24 the claim is only 25\% (phones) and 21\% (embodied) present, and a logged
web click 12\%. By layer 30 these rise to 62\%, 46\% and 61\%.
\item \emph{Nothing is decided before layer 16:} patching earlier has no effect.
\end{itemize}
The larger decider-12b (48 layers) shows the same order in all three domains: by layer 32, patching the observation restores
46--66\% of its effect, while the logged step restores 10--42\% and the claim 22--38\%; the record reaches 63--85\% only by
layer 42 (Appendix Table~\ref{tab:patch12b}).
Record-dependent influences thus become increasingly dominant in later layers, even though outcome information is already
available in the intermediate representation at the decision position. The
record does not travel alone, however. Removing the record's own tokens from the last layers (and keeping the rest of the
sequence) leaves the decision unchanged, because by then the record has spread into the tokens that follow it: later pages,
the question and the options (Appendix~\ref{app:blind}). This is why reading the middle layer works better than editing the
late computation: on a synthetic diagnostic set, steering the answer position raised accuracy on claimed-but-unfinished
states from 0.48 to 0.70 at best, while simply reading the middle-layer probe reached 0.89.

\section{\sysname{}: Detecting the Mismatch}
\label{sec:detector}
\begin{figure}[t]
\centering
\definecolor{evc}{HTML}{2A78D6}\definecolor{clc}{HTML}{EB6834}\definecolor{inkc}{HTML}{52514E}
\begin{tikzpicture}[font=\tiny, >=Stealth, line width=0.5pt,
  box/.style={draw=inkc, rounded corners=2pt, inner sep=2.5pt, align=left}]
\node[box, anchor=west] (x) at (0,0) {{\scriptsize\bfseries Agent state}\\[2pt]
  goal: turn on Wi-Fi\\[1pt]
  record: \textcolor{clc}{``Wi-Fi}\\ \textcolor{clc}{\phantom{record: }is now on.''}\\[1pt]
  screen: \textcolor{evc}{toggle}\\ \textcolor{evc}{\phantom{screen: }still off}};
\path (x.east) ++(0.32,0) coordinate (m);
\foreach \i in {0,...,11} {
  \ifnum\i=7 \def\cc{evc}\else\ifnum\i>8 \def\cc{clc!45}\else\def\cc{inkc!25}\fi\fi
  \fill[\cc] ($(m)+(0.2*\i,-0.42)$) rectangle ++(0.13,0.84);}
\node[anchor=south, text=inkc] at ($(m)+(1.165,0.44)$) {decision model (one pass)};
\node[anchor=north, text=clc, align=center] at ($(m)+(2.36,-0.45)$) {record\\dominates};
\node[anchor=north east, text=evc] at ($(m)+(1.39,-0.45)$) {layer $\ell$};
\draw[->, inkc] (x.east) -- ($(m)+(-0.06,0)$);
\node[box, anchor=west] (head) at ($(m)+(2.75,0)$) {{\scriptsize\bfseries Output}\\[2pt] $p(\textsf{stop})=0.91$\\[1pt] proposes \textsf{stop}};
\draw[->, inkc] ($(m)+(2.36,0)$) -- (head.west);
\node[box, draw=evc, fill=evc!7, line width=0.8pt, anchor=north] (g) at ($(m)+(1.465,-1.2)$)
  {{\scriptsize\bfseries\textcolor{evc}{EviGate}}\\[2pt] $s=\sigma(w^{\top}\tilde h_\ell)$\\[1pt] $\phantom{s}=0.08<\tau$\\[1pt] step did not work};
\draw[->, evc, line width=0.8pt] ($(m)+(1.465,-0.45)$) -- (g.north);
\node[box, anchor=north west] (dec) at (head.west |- g.north) {{\scriptsize\bfseries Veto}\\[2pt] terminal op, $s<\tau$\\[1pt] $\Rightarrow$ \textsf{continue}};
\draw[->, inkc] (head.south -| dec.north) -- (dec.north);
\draw[->, evc, line width=0.8pt] (g.east) -- (g.east -| dec.west);
\end{tikzpicture}
\caption{\sysname{} on an illustrative phone step. The record says Wi-Fi is on; the screen shows it is off. The output follows
the record and proposes \textsf{stop}. The gate reads the decision-position state at a middle layer $\ell$ of the
\emph{same} forward pass, before record-dependent influence dominates the last layers, and vetoes. It acts only on terminal options.
Probabilities are illustrative.}
\label{fig:system}
\end{figure}
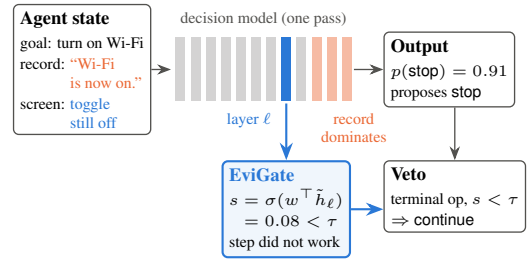

\paragraph{The detector.} Let $h_\ell$ be the decision model's hidden state at layer $\ell$ and at the position where it
reads its answer; the forward pass computes it anyway. \sysname{} keeps a linear readout $s = \sigma(w^\top \tilde h_\ell +
b)$ on the standardised state, an estimate of the probability that the last step took effect. When the model proposes a
terminal option and $s<\tau$, the gate vetoes it and the agent takes the best non-terminal option instead
(Algorithm~\ref{alg:gate}, Figure~\ref{fig:system}). It needs no extra forward pass and adds 0.04--2.0\% latency in
decider's production engine and Kev's serving path, against 56--98\% for a second question or contrastive decoding
(Appendix~\ref{app:serving}).

\begin{algorithm}[t]
\caption{\sysname{} at one agent step}
\label{alg:gate}
\begin{algorithmic}
\STATE {\bfseries Input:} state $x$, question $q$, readout $(w,b,\ell)$, threshold $\tau$
\STATE $p, h_\ell \leftarrow \textsc{DecisionModel}(x, q)$ \hfill\COMMENT{one pass}
\STATE $a \leftarrow \arg\max_o p(o)$
\IF{$a$ is terminal (stop, proceed, buy)}
  \STATE $s \leftarrow \sigma(w^\top \tilde h_\ell + b)$ \hfill\COMMENT{$O(d)$}
  \IF{$s < \tau$}
     \STATE $a \leftarrow \arg\max_{o \notin \text{terminal}} p(o)$ \hfill\COMMENT{veto}
  \ENDIF
\ENDIF
\STATE {\bfseries return} $a$
\end{algorithmic}
\end{algorithm}

\paragraph{Fitting it.} The readout $(w, b, \ell)$ is fit on states labelled by whether the step took effect. The labels come
free from the environment: an option is selected or not, an action returned ``Nothing happens.'' or not.

\paragraph{Choosing the threshold.} $\tau$ trades two errors: letting a failed step through, and vetoing a step that worked.
We choose the $\tau$ that maximises balanced accuracy on held-out calibration states.

\subsection{Offline evaluation}
\label{sec:offline}
\begin{table}[t]
\caption{Gating terminal decisions on held-out groups: balanced accuracy (0.5 is chance). \emph{Own score}: the model's
own probability of the terminal option, with a tuned threshold. \emph{Second q.}: an added yes/no question asking whether the
step worked. \emph{Wrong}: failed steps that still end in a terminal decision, with the own score $\to$ with \sysname{}.}
\label{tab:offline3}
\centering\small
\setlength{\tabcolsep}{2pt}
\begin{tabular}{@{}lccc@{\hskip 8pt}c@{}}
\toprule
Model & Own score & Second q. & \sysname{} & Wrong \\
\midrule
\multicolumn{5}{@{}l}{\emph{Phones}} \\
decider-4b & .59 & .60 & $\mathbf{.71}$ & 72\% $\to$ 38\% \\
Kev-4B & .60 & .61 & $\mathbf{.72}$ & 58\% $\to$ 38\% \\
decider-12b & .63 & .62 & $\mathbf{.73}$ & 33\% $\to$ 33\% \\
hosted Jev & .66 & .69 & -- & -- \\
\midrule
\multicolumn{5}{@{}l}{\emph{Embodied (ALFWorld)}} \\
decider-4b & .84 & .97 & $\mathbf{1.00}$ & 18\% $\to$ 0\% \\
Kev-4B & .69 & .81 & $\mathbf{.99}$ & 21\% $\to$ 0.5\% \\
decider-12b & .65 & .75 & $\mathbf{1.00}$ & 36\% $\to$ 0\% \\
hosted Jev & .78 & .99 & -- & -- \\
\midrule
\multicolumn{5}{@{}l}{\emph{Web (WebShop)}} \\
decider-4b & .56 & .52 & $\mathbf{1.00}$ & 68\% $\to$ 0\% \\
Kev-4B & .68 & .66 & $\mathbf{1.00}$ & 38\% $\to$ 0.2\% \\
decider-12b & .54 & .50 & $\mathbf{1.00}$ & 60\% $\to$ 0\% \\
hosted Jev & .53 & .76 & -- & -- \\
\bottomrule
\end{tabular}
\end{table}

Table~\ref{tab:offline3} compares \sysname{} with the two detectors that need no training. Retuning the threshold on the
model's own score helps little, because the record has already corrupted its ranking. Asking a second yes/no question
inherits the same bias on phones and web and works only with ALFWorld's explicit failure phrase. \sysname{} is the best
detector for every model and domain: it gains 0.10--0.12 balanced accuracy on phones and is near-perfect in ALFWorld and
WebShop. On phones, the same holds for all seven models and for contrastive decoding~\cite{shi2024cad} (Appendix
Table~\ref{tab:stop}). Simply deleting the record from the input is no substitute: the record carries progress
information, so on phones the record-free model vetoes 93\% of correct completions, and in ALFWorld and WebShop it still
passes 6--10\% of failed steps, against at most 0.5\% for \sysname{} (Appendix~\ref{app:blind}).

\section{Detecting Mismatches in Live Agents}
\label{sec:loops}
Offline accuracy does not show that an agent finishes more tasks: a veto helps only if the agent then does something
better. We therefore run full agents. Every condition uses the same tasks and random seeds, so differences are paired.

\paragraph{Agents.}
\begin{itemize}
\item \emph{Phones (simulator).} Live AndroidWorld needs KVM emulation, which we did not have, so we use a text phone
simulator with 24 tasks in 11 apps, 30 seeds per task. The decisive tap fails silently 40\% of the time, and the simulator's record generator
still claims success after 70\% of failed taps.
\item \emph{ALFWorld and WebShop.} As in the ReAct-J agent of~\citet{ling2026faster}, an LLM (Qwen3.5-9B in ALFWorld,
Qwen3.5-4B in WebShop) writes a thought before each action, and decider-4b chooses the action. The records are these
thoughts. We use 134 unseen ALFWorld tasks and 200 WebShop goals, with one run per task (greedy thoughts and a fixed
failure seed), so their confidence intervals reflect variation across tasks. The gate checks moving on after a
state-changing action (ALFWorld; a veto retries the action) and \texttt{buy now} (WebShop).
\end{itemize}
To make silent failures frequent enough to measure, actions and option clicks fail silently 30\% of the time, a stress test
in the spirit of fault-injection benchmarks for web agents~\cite{bai2026stressweb}. Without injection, the mismatch still
arises from the agent's own optimistic records (\S\ref{sec:measure}).

\paragraph{Conditions} (Table~\ref{tab:loops3}). \emph{No gate}: the agent takes whatever the decision model chooses.
\emph{Second question}: before accepting a terminal decision, ask the decision model whether the last step worked (one extra
call). \emph{Random veto}: veto terminal decisions at random, as often as the gate it is compared with; any gain over it
comes from \emph{which} decisions a gate vetoes, not from vetoing more. \sysname{} is fit either on \emph{offline contrast
sets}, the paired states of \S\ref{sec:setup}, or on \emph{target-domain states}: states recorded from the same agent on other tasks (ALFWorld:
148 states from 100 training games; WebShop: 480 states from 300 other goals).

\begin{table*}[t]
\caption{Task success (\%) of live agents; same tasks and seeds in every row. Random veto: vetoes at random, about as often
as the \sysname{} row above it (rates in the note below). Bottom: paired differences with 95\% CIs over tasks.}
\label{tab:loops3}
\centering\small
\setlength{\tabcolsep}{10pt}
\begin{tabular}{@{}lccc@{}}
\toprule
Condition & Phones$^*$ & ALFWorld & WebShop \\
\midrule
No gate & 77.2 & 42.5 & 8.5 \\
Second question & $\mathbf{83.2}$ & $\mathbf{61.2}$ & 7.0 \\
\sysname{}, fit on offline contrast sets & -- & 58.2 & 2.5 \\
\midrule
\sysname{}, fit on target-domain states & 82.4 & $\mathbf{61.2}$ & $\mathbf{13.5}$ \\
\quad random veto, matched rate & 77.6 & 58.2 & 4.5 \\
\midrule
\multicolumn{4}{@{}l}{\emph{\sysname{} $-$ no gate}} \\
& $+5.1$ {\scriptsize[1.8, 8.9]} & $+18.7$ {\scriptsize[11.9, 26.1]} & $+5.0$ {\scriptsize[1.0, 9.5]} \\
\multicolumn{4}{@{}l}{\emph{\sysname{} $-$ random veto}} \\
& $+4.7$ {\scriptsize[1.7, 8.2]} & $+3.0$ {\scriptsize[$-$1.5, 7.5]} & $+9.0$ {\scriptsize[4.5, 13.5]} \\
\bottomrule
\end{tabular}
\vskip 3pt\parbox{0.8\textwidth}{\raggedright\footnotesize $^*$Simulator. Target-domain states are simulator-generated
template states from the other half of the tasks; in ALFWorld and WebShop they are states recorded from the same agent on
separate training tasks. The gate fit on offline contrast sets was not run in the simulator. Random veto rates: phones,
the gate's measured rate (21.8\% and 9.1\% vs.\ 21.8\% and 8.0\% for \sysname{} on the two task halves); WebShop, the gate's
measured rate (63.6\% vs.\ 67.1\%); ALFWorld, the failure rate among gated steps on the training games (78.6\% vs.\ 75.5\%).}
\end{table*}

\paragraph{Results.}
\begin{itemize}
\item \emph{Detection raises task success in every domain.} \sysname{} adds 5.1 points on the
phone simulator, 18.7 in ALFWorld and 5.0 in WebShop. On phones and the web it also beats random vetoing at the same rate
($+4.7$, $+9.0$), so the gain comes from vetoing the right decisions. On the simulator it matches a 9B LLM judge without the
judge's extra calls (Appendix~\ref{app:loops}).
\item \emph{A second question is a strong baseline where failure is easy to see.} In the simulator and in ALFWorld, whose
failures are templated or stated explicitly, asking the decision model whether the step worked does as well as \sysname{}
(83.2 vs.\ 82.4; 61.2 vs.\ 61.2), at the cost of an extra question per decision. In WebShop it fails (7.0, below no gate).
\sysname{} is the only method that helps in all three domains.
\item \emph{The gate must be fit on target-domain states.} Fit on offline contrast sets, the gate fails in real agents: in WebShop it
vetoes 95\% of purchases and success falls to 2.5\%; in ALFWorld it is no better than random. Live records look different
from offline ones (a thought at every step, other page sequences), and a few hundred states from the agent itself fix it.
\item \emph{Recovery limits the gain.} After a vetoed purchase the WebShop agent often wanders instead of selecting the
missing option, so it buys less (46 vs.\ 78 purchases). In ALFWorld, 80\% of gated steps follow a real failure, so retrying
is almost always right and random vetoing recovers most of the gain. A detector is most valuable where vetoes are rare and
must be precise, and only if the agent has a good next action.
\end{itemize}
The same readout can also pick the best of several attempts: in best-of-16 sampling on WebShop it chooses a successful
attempt 12.0 points more often than the decision model's own judgement (Appendix~\ref{app:bon}).

\section{Why Not Fine-Tune the Model?}
\label{sec:finetune}
If the model knows, could we simply train its output to use what it knows? This is the standard remedy when a model trusts
the wrong input: counterfactual data augmentation~\cite{kaushik2020cad}, training on conflicting contexts so a model follows
its context over its memory~\cite{li2023kaft}, synthetic data against sycophancy~\cite{wei2023sycophancy}, and adversarial
negatives for reward models~\cite{zhao2025onetoken}. We apply it to decider-4b: a LoRA adapter (rank 16, 0.8\% of the
parameters) trained on paired states with one label, whether the step took effect. A second loss term keeps the model's
preferences among the non-terminal options unchanged (details in Appendix~\ref{app:ft}). We train three adapters: on phones only (2{,}755
states from 205 of the 245 apps); on all three domains with 1{,}000 states each (from training apps, games and goals); and
on all three domains with all 2{,}755 phone states. All are tested on held-out apps, games and goals.

\begin{table}[t]
\caption{Fine-tuning decider-4b on paired states. \emph{Wrong}: failed steps whose record claims success that still end in a
terminal decision. \emph{AUROC}: the output's detection of the failure on claimed states. Tests use held-out apps,
games and goals. Last rows: accuracy on two of
decider's held-out general tasks.}
\label{tab:ft}
\centering\footnotesize
\setlength{\tabcolsep}{2pt}
\begin{tabular}{@{}llcccc@{}}
\toprule
& & & \multicolumn{3}{c}{Fine-tuned on} \\
\cmidrule(l){4-6}
Test domain & & Base & phones & all 3 & all 3 \\
 & & & & 1k each & all phones \\
\midrule
Phones & Wrong & 91\% & $\mathbf{47\%}$ & 89\% & $\mathbf{47\%}$ \\
 & AUROC & .50 & $\mathbf{.84}$ & .67 & .82 \\
ALFWorld & Wrong & 90\% & 91\% & $\mathbf{0\%}$ & 0.3\% \\
 & AUROC & .91 & .92 & $\mathbf{1.00}$ & $\mathbf{1.00}$ \\
WebShop & Wrong & 100\% & 100\% & $\mathbf{0\%}$ & $\mathbf{0\%}$ \\
 & AUROC & .46 & .48 & $\mathbf{1.00}$ & $\mathbf{1.00}$ \\
\midrule
HWU64 & Acc. & 84.5 & 84.3 & 84.8 & 85.0 \\
DBpedia & Acc. & 77.5 & 78.3 & 78.3 & 78.5 \\
\bottomrule
\end{tabular}
\end{table}

Table~\ref{tab:ft} gives three findings.
\begin{itemize}
\item \emph{In-domain, fine-tuning works.} Trained on phones, the model stops after 47\% instead of 91\% of claimed failures
on unseen apps. The screen now drives the decision (its effect rises from $+0.02$ to $+1.34$) and the claim barely matters
($+0.74 \to +0.08$). On claimed states the output's AUROC (.84) even exceeds the base model's probe (.74). General accuracy is unchanged.
\item \emph{It does not transfer across domains.} The phone-trained model is unchanged in ALFWorld and WebShop. What it
learns is how failure looks on a phone screen, not a general rule to trust the observation over the record.
\item \emph{With enough data from every domain, one model fixes all three.} Trained on 1{,}000 states from each domain,
the model makes no wrong decisions in ALFWorld and WebShop but loses most of the phone gain (89\%). With all 2{,}755
phone states plus the same ALFWorld and WebShop states, it matches the phone-only model on phones (47\%, AUROC .82) while
staying near-perfect in ALFWorld (0.3\%) and WebShop (0\%), with general accuracy unchanged. Training on several domains
therefore need not cost accuracy, as long as each domain has enough paired states.
\end{itemize}
Fine-tuning and detection therefore need the same thing, labelled states from the target domain, but differ in what they
change: fine-tuning edits a model shared across every task it serves and needs thousands of paired states per domain, while
\sysname{} leaves the model untouched and needs a few hundred labelled states from the deployment. Where the model's
weights cannot be changed, as for hosted Jev, neither applies directly, which motivates exposing intermediate states in
hosted decision APIs. We evaluate fine-tuning offline only; whether its gains carry over to live agents, as \sysname{}'s
do, remains to be tested.

\section{Related Work}

\paragraph{Decision models in agents.}
Decision models answer typed questions in a single forward pass, making them faster and cheaper than LLM generation~\cite{typesafe2026jev,marosi2026decider,palmer2026kev}. Thousands of public projects use Jev for routing, scoring, and action selection~\cite{ling2026jevwild}, and replacing LLM-based action selection with Jev reduces per-decision costs~\cite{ling2026faster}. Prior work has identified several failure modes: decision models favor option names over their definitions~\cite{azizi2026labels,wording2026}, adversarially renamed options can bypass guardrails~\cite{azizi2026option}, and premature completion decisions reduce end-to-end task success~\cite{ling2026faster}. We study a different failure mode that requires no attacker: a decision model trusts the agent's record of its actions over conflicting evidence from the environment.

\paragraph{Unreliable agent records.}
LLM agents often report success without corresponding evidence in the environment~\cite{falsesuccess2026,progressbar2026}. Such inaccurate records can mislead downstream components. For example, safety monitors may approve risky actions when an upstream summary omits an ``unverified'' marker~\cite{hu2026silence}. We examine how such discrepancies affect downstream decision models, and why these models favor the records.

\paragraph{Probing and gating agents.}
Prior work shows that probes trained on agent hidden states can predict task success~\cite{agentsknow2026} and episode failure~\cite{doomed2026}. In agent safety, \citet{fayyazi2026agentknew} find that false statements often precede unauthorized actions, while lies follow them. They use hidden-state probes to predict and gate these actions, and steering to reduce subsequent lies. Unlike these approaches, which examine the agent's own hidden states, we study decision models that consume the agent's records. Using activation patching, we identify where these models begin to favor agent records over conflicting environmental observations. We then show that evidence of the conflict remains accessible in intermediate layers and can be used to detect mismatches. Finally, we show that the detector must be calibrated on states produced by the agent itself.

\section{Limitations}
To obtain enough mismatches for evaluation, our live agents run with injected silent failures (30\% of actions in ALFWorld and WebShop; in the phone simulator, 40\% of first attempts and 20\% of retries). Our offline experiments show the same behavior without injecting failures into live execution, using real trajectories and LLM-written summaries (\S\ref{sec:measure}); how often such mismatches arise naturally in deployment remains unknown. Because Android emulation was unavailable, our live phone agent runs in a text simulator, and real phone screens are evaluated offline only. In ALFWorld and WebShop, failed actions leave explicit textual traces that give reliable labels but make mismatches easier to detect; the phone domain, built from real screens where failures are implicit, covers the harder case. Finally, our mitigation methods require labelled states from the target agent and access to model internals: \sysname{} relies on hidden activations, while fine-tuning additionally requires model weights. The hosted Jev model shows the same bias but can only be measured.

\section{Conclusion}
Agents record what they did, and the fast decision models that act on those records trust them over the world: on phones,
in embodied tasks and in web shopping, in open models from 0.8B to 12B parameters and in hosted Jev. The models are not blind: the observation is
present in their middle layers, but record-dependent influences dominate the final layers. Reading the middle layer detects
the mismatch with one dot product and raises task success in all three domains when calibrated on the agent's own states.
Fine-tuning fixes the output too, but only in the domains it is trained on. For builders of agents, the practical advice is
to keep unverified records out of the decision state where possible, and otherwise to check terminal decisions against what
the model already encodes.

\bibliography{refs}
\bibliographystyle{mlsys2025}


\appendix
\onecolumn
\paragraph{Appendix overview.}
Appendix~\ref{app:details} specifies data construction, model interfaces, probing, agent execution, and compute.
Appendix~\ref{app:phones} reports complete phone-domain results and additional baselines supporting
\S\ref{sec:measure} and \S\ref{sec:latent}. Appendix~\ref{app:blind} tests whether removing record tokens
from later layers changes decisions (\S\ref{sec:why}). Appendix~\ref{app:serving} reports the runtime overhead of
\sysname{} (\S\ref{sec:detector}).
Appendix~\ref{app:loops} gives the complete phone-simulator comparison (\S\ref{sec:loops});
Appendix~\ref{app:ft} details fine-tuning (\S\ref{sec:finetune}); and Appendix~\ref{app:bon}
evaluates the same probe as a best-of-$N$ verifier.

\section{Experimental Setup}
\label{app:details}

\paragraph{Phone decision requests.}
We use the T3A decision interface of~\citet{ling2026faster} unchanged. Each request has three parts.
\begin{itemize}
\item \emph{State:} T3A's action-selection prompt $P$, filled with the goal, the summaries of the earlier steps and the
elements of the current screen, and wrapped as \texttt{\{"native\_t3a\_prompt": $P$\}}. The template is
\texttt{ACTION\_SELECTION\_PROMPT\_TEMPLATE} in \texttt{android\_world/agents/t3a.py}.
\item \emph{Question:} a typed \texttt{choice}, \emph{``Following native\_t3a\_prompt, choose the next single action type to
advance the goal.''}
\item \emph{Options:} twelve action types, namely interaction (\texttt{click}, \texttt{long\_press}, \texttt{input\_text},
\texttt{scroll}, \texttt{keyboard\_enter}), navigation (\texttt{open\_app}, \texttt{navigate\_back},
\texttt{navigate\_home}), \texttt{wait} and \texttt{answer}, and the two terminal operations \texttt{complete} (``The
requested goal is already achieved.'') and \texttt{infeasible}.
\end{itemize}
The second-question baseline adds one yes/no question to the same request: \emph{``Judging from the UI elements of the
current screen in native\_t3a\_prompt, has the user goal already been achieved?''}

\paragraph{LLM-written summaries.}
For each of the 200 episodes of the 1{,}000-state phone subset, Qwen3.5-9B rewrites the final step's summary with T3A's
step-summary prompt (\texttt{SUMMARIZATION\_PROMPT\_TEMPLATE} in \texttt{android\_world/agents/t3a.py}), given the goal, the
action, and the screen before and after the step; in the failure condition the after-screen equals the before-screen.
Generation is greedy (no sampling, thinking disabled, at most 120 new tokens; the first line is kept). A summary counts as
claiming success if the same model, asked ``Does the summary state or imply that the action succeeded or that the goal is
now accomplished? Answer yes or no.'', answers yes; 78 of the 200 failed-step summaries do. This automatic judgement was not
verified by human annotation.

\paragraph{Phone screens and action records.}
We extract screen elements with T3A's \texttt{forest\_to\_ui\_elements} and validity filter.
Each element is represented as a JSON object containing its nonempty fields, including text,
content description, hint, class, resource name, checked and selected states, and interaction flags
(clickable, editable, scrollable). We exclude screen descriptions longer than 9{,}000 characters.
From AndroidControl shards 0--1, we retain episodes containing at least two actions, with a final
action that visibly changes the screen, is neither navigation nor \texttt{wait}, and does not
answer a question-type goal. For a recorded action $i$, the logged-step template is
``Step $i$: \emph{Verb}[ text] in order to \emph{instruction}.'' An explicit success claim is
added using one of four fixed sentences, such as ``This completes the requested task.''
The offline contrasts use actual before- and after-action screen descriptions, but the failed-action
variant is constructed by pairing the action record with the unchanged earlier screen; it is not
an independently observed failed Android execution.

\paragraph{ALFWorld and WebShop contrast sets.}
For ALFWorld, we run the environment's hand-coded expert on all 134 \texttt{valid\_unseen} tasks;
117 are solved. We form five-state contrast sets at up to three state-changing expert steps per
task. The claim is a ReAct-style thought:
``think: Now I have done `\emph{a}'. Next, I need to \emph{a$'$}.''
The candidate actions are the expert's next command, a retry of the previous command, and two
other valid commands. For WebShop, we use the 1{,}000-product catalogue with attribute-based
goals and BM25 (\texttt{rank\_bm25}) search over WebShop's indexed product text. For each goal,
we open the target product and execute the required option selections in the environment.
A failed variant retains the recorded clicks but presents the page with those options unselected.
The four candidate operations are \texttt{buy now}, the first required option,
\texttt{description}, and \texttt{back to search}.

\paragraph{Reading decision models.}
For decider, we take the answer-position hidden state (at ``Answer: ('') and use the released
scoring temperatures. For Kev, we use its \texttt{<decide>} representation and pointer head,
with LoRA weights merged. For \texttt{laya-multilingual}, we average the \texttt{[CLS]}
representations across input windows. We evaluate the Qwen3.5-4B base and chat models with
decider's prompts and option-letter scoring, without additional training.

\paragraph{Probes and uncertainty estimates.}
We extract representations from every second layer, standardize them using training-fold
statistics, and fit $\ell_2$-regularized logistic regression with $C=0.02$.
We use five-fold GroupKFold splits by app, ALFWorld task, or WebShop goal, so variants of the
same evaluation group cannot appear in both training and testing. Within each training fold,
a four-fold grouped cross-validation chooses the layer from the upper two-thirds of model depth.
We estimate paired log-probability effects using 1{,}000 bootstrap resamples at the episode level;
AUROC differences use 1{,}000 paired bootstrap resamples of apps. The phone text-only baseline
is logistic regression on TF--IDF features from screen text, using word unigrams and bigrams.

\paragraph{Live ALFWorld agent.}
The state consists of a ReAct transcript with two in-context examples.
Before each action, Qwen3.5-9B generates a thought greedily (at most 60 tokens), and decider-4b
selects from valid commands using the ALFWorld instruction of~\citet{ling2026faster}.
Each state-changing command independently has a 0.3 probability of injected failure: the
environment does not execute the command and returns ``Nothing happens.''
The agent has at most 50 steps. We fit the gate on 148 labelled decision states collected from
100 separate ALFWorld \texttt{train}-split games, then evaluate on the unseen-task split.

\paragraph{Live WebShop agent.}
Qwen3.5-4B writes the search query and one thought for each page, while decider-4b chooses
among clickable elements. Option clicks have a 0.3 probability of silently doing nothing;
the episode ends after at most 15 steps. For gate training, we collect 480 states from the
agent on WebShop goals 1000--1299; the label for purchase verification indicates whether all
required options are selected. The gate acts at purchase decisions, not at every page visit.
For both environments, the probe reads layer 20. The balanced thresholds are 0.158 for ALFWorld
and 0.229 for WebShop.
Random-veto controls in WebShop and the phone simulator veto at the measured rate of the gate they are compared with; in
ALFWorld, at the failure rate among gated steps on the training games (79.7\% targeted, 78.6\% measured, vs.\ 75.5\% for
\sysname{}).

\paragraph{Phone simulator.}
We simulate 24 tasks spanning 11 apps. Each task starts at a home screen and requires opening
the correct app and performing one state-changing action. The available options include two
distractors, \emph{Navigate back}, and \emph{Stop}; screens contain four to six distractor
elements, and option order is randomized. The decisive action does nothing with probability
0.4 on its first attempt and 0.2 on later attempts. After a failure, the simulated action
record falsely states ``\emph{goal} -- done.'' with probability 0.7; otherwise it reports
``The screen did not change.'' The step limit is eight. This phone experiment is a text
simulator with a stochastic record generator, not a live Android emulator or an LLM-written
summary at each step. The gate reads layer 20 and is fitted on 384 template states from the
other half of the task set.

\paragraph{Compute.}
Experiments use two NVIDIA RTX A6000 GPUs (48\,GB each), bf16 computation, PyTorch 2.14,
and Transformers 5.19. Approximate compute is 20 GPU-hours for offline measurements,
60 GPU-hours for live-agent experiments (including best-of-$N$), and 5 GPU-hours for
fine-tuning.

\section{Uncertainty of the Input Effects}
\label{app:ci}
Table~\ref{tab:effectsci} gives paired bootstrap confidence intervals (2{,}000 resamples of decision points) for the
comparisons behind Table~\ref{tab:effects3}. \emph{Logged $-$ Obs.} and \emph{Claim $-$ Obs.} are within-point differences
between a record effect and the observation effect; \emph{Obs.\ $\mid$ claim} is the effect of changing the observation from
failed to successful while the false success claim stays in the record (\textsc{Done+claim} $-$ \textsc{Logged+claim}). In
every configuration at least one record effect exceeds the observation effect with an interval above zero. With the claim
present, the observation has little influence on the decision.

\begin{table}[h]
\caption{Paired 95\% bootstrap intervals for the effects in Table~\ref{tab:effects3} (differences in $\log p$ of the terminal
option). Decision points: phones 640 (decider-4b, Kev-4B) or 200 (decider-12b, Jev); ALFWorld 295; WebShop 300.}
\label{tab:effectsci}
\centering\small
\setlength{\tabcolsep}{6pt}
\begin{tabular}{@{}lccc@{}}
\toprule
Model & Logged $-$ Obs. & Claim $-$ Obs. & Obs.\ $\mid$ claim \\
\midrule
\multicolumn{4}{@{}l}{\emph{Phones}} \\
decider-4b & $+1.05$ {\scriptsize[0.94, 1.17]} & $+0.74$ {\scriptsize[0.69, 0.80]} & $+0.00$ {\scriptsize[$-$0.01, 0.01]} \\
Kev-4B & $+0.11$ {\scriptsize[0.09, 0.14]} & $+0.23$ {\scriptsize[0.22, 0.25]} & $+0.05$ {\scriptsize[0.04, 0.06]} \\
decider-12b & $+0.89$ {\scriptsize[0.68, 1.09]} & $+0.34$ {\scriptsize[0.23, 0.46]} & $+0.03$ {\scriptsize[$-$0.02, 0.09]} \\
hosted Jev & $+1.70$ {\scriptsize[1.33, 2.10]} & $+0.34$ {\scriptsize[0.22, 0.47]} & $+0.14$ {\scriptsize[0.11, 0.18]} \\
\midrule
\multicolumn{4}{@{}l}{\emph{ALFWorld}} \\
decider-4b & $+1.41$ {\scriptsize[1.18, 1.65]} & $+0.09$ {\scriptsize[0.00, 0.19]} & $+0.24$ {\scriptsize[0.18, 0.29]} \\
Kev-4B & $-0.03$ {\scriptsize[$-$0.05, $-$0.00]} & $+0.04$ {\scriptsize[0.01, 0.07]} & $+0.23$ {\scriptsize[0.21, 0.25]} \\
decider-12b & $+0.80$ {\scriptsize[0.68, 0.93]} & $+0.50$ {\scriptsize[0.42, 0.58]} & $+0.12$ {\scriptsize[0.10, 0.13]} \\
hosted Jev & $+3.42$ {\scriptsize[3.06, 3.78]} & $+0.18$ {\scriptsize[0.09, 0.28]} & $+0.22$ {\scriptsize[0.19, 0.25]} \\
\midrule
\multicolumn{4}{@{}l}{\emph{WebShop}} \\
decider-4b & $+3.69$ {\scriptsize[3.58, 3.81]} & $+0.05$ {\scriptsize[0.04, 0.06]} & $-0.00$ {\scriptsize[$-$0.00, $-$0.00]} \\
Kev-4B & $+0.20$ {\scriptsize[0.18, 0.22]} & $+0.12$ {\scriptsize[0.11, 0.14]} & $+0.06$ {\scriptsize[0.05, 0.07]} \\
decider-12b & $+1.29$ {\scriptsize[1.18, 1.39]} & $+1.07$ {\scriptsize[0.95, 1.18]} & $+0.01$ {\scriptsize[$-$0.01, 0.02]} \\
hosted Jev & $+2.54$ {\scriptsize[2.40, 2.69]} & $+0.37$ {\scriptsize[0.32, 0.42]} & $+0.02$ {\scriptsize[0.01, 0.02]} \\
\bottomrule
\end{tabular}
\end{table}

\section{Additional Phone Results}
\label{app:phones}

The main text compares the four models evaluated in all three domains. Here we report all
seven open models evaluated on phones, varying model size (Kev-0.8B, 4B, and 9B;
decider-2b, 4b, and 12b) and architecture (the encoder-based Laya).
Unless otherwise marked, results use the complete set of 3{,}200 counterfactual states.
Rows marked $^\dagger$ use a 1{,}000-state subset.

\paragraph{Sensitivity to observations and records.}
Table~\ref{tab:effects} reports the effect of changing the observed outcome, the logged action,
or the explicit success claim, each with other inputs fixed. The effect of adding a success
claim exceeds the observation effect for every model up to 4B. This pattern does not hold
uniformly at larger scales: Kev-9B gives greater weight to the screen than to either record
component considered separately, whereas decider-12b remains most sensitive to the action
log. The evidence therefore supports a widespread, but not scale-invariant, preference for
record cues.

\begin{table}[h]
\caption{Paired effects on $\log p(\texttt{complete})$ on phones, with 95\% bootstrap confidence intervals. For models up to 4B, explicit claims exert a larger mean effect than screen evidence. The larger models differ: Kev-9B weights the observed outcome more than either record component individually, whereas decider-12b remains most sensitive to the logged action. Rows marked $^\dagger$ use a 1{,}000-state subset.}
\label{tab:effects}
\centering\small
\begin{tabular}{@{}lccc@{}}
\toprule
Model & Observation & Logged step & Claim \\
\midrule
decider-4b & $+0.05$ {\scriptsize[.02,.08]} & $\mathbf{+1.10}$ {\scriptsize[1.00,1.21]} & $+0.80$ {\scriptsize[.74,.85]} \\
decider-2b$^\dagger$ & $+0.02$ {\scriptsize[$-$.02,.05]} & $+0.48$ {\scriptsize[.38,.58]} & $\mathbf{+0.79}$ {\scriptsize[.72,.87]} \\
Kev-4B & $+0.09$ {\scriptsize[.07,.10]} & $+0.20$ {\scriptsize[.18,.22]} & $\mathbf{+0.32}$ {\scriptsize[.31,.33]} \\
Kev-0.8B$^\dagger$ & $+0.00$ {\scriptsize[.00,.01]} & $+0.02$ {\scriptsize[.01,.02]} & $\mathbf{+0.21}$ {\scriptsize[.20,.22]} \\
Laya$^\dagger$ & $-0.04$ {\scriptsize[$-$.13,.06]} & $-0.01$ {\scriptsize[$-$.06,.04]} & $\mathbf{+0.32}$ {\scriptsize[.27,.38]} \\
\midrule
Kev-9B$^\dagger$ & $\mathbf{+0.34}$ {\scriptsize[.28,.41]} & $+0.27$ {\scriptsize[.22,.32]} & $+0.24$ {\scriptsize[.22,.27]} \\
decider-12b$^\dagger$ & $+0.21$ {\scriptsize[.11,.32]} & $\mathbf{+1.10}$ {\scriptsize[.93,1.27]} & $+0.56$ {\scriptsize[.47,.66]} \\
\bottomrule
\end{tabular}
\end{table}

\paragraph{Output scores versus internal probes.}
Table~\ref{tab:probe} gives AUROC for detecting whether the final phone action took effect.
The \emph{Claimed} columns contain both successful and failed states with success-claim
language, requiring discrimination from other evidence. For decider-4b and Kev-4B, claimed-state
AUROC rises from 0.51 and 0.55 at the output to 0.74 and 0.79 with a linear probe.
The corresponding all-state gains are 0.14 and 0.15, with positive bootstrap confidence
intervals. The smallest Kev model shows a smaller gain (0.04, 95\% CI $[-0.01,0.09]$),
so improvement is not statistically established for every model. The input-text classifier
reaches 0.59 AUROC, below the probe results for the 4B models.

\begin{table}[h]
\caption{Detecting whether the phone step worked (AUROC on held-out apps), as in Table~\ref{tab:probe3}. \emph{Gap}:
probe minus output on all states, with 95\% CI. \emph{Text only}: a classifier on the screen text alone.}
\label{tab:probe}
\centering\small
\setlength{\tabcolsep}{4pt}
\begin{tabular}{@{}lcccccc@{}}
\toprule
& \multicolumn{2}{c}{Output} & \multicolumn{2}{c}{Probe} & & \\
\cmidrule(lr){2-3}\cmidrule(lr){4-5}
Model & All & Claimed & All & Claimed & Gap & Text only \\
\midrule
decider-4b & .63 & .51 & $\mathbf{.77}$ & $\mathbf{.74}$ & $+.14$ {\scriptsize[.11,.17]} & .59 \\
Kev-4B & .65 & .55 & $\mathbf{.80}$ & $\mathbf{.79}$ & $+.15$ {\scriptsize[.12,.17]} & .59 \\
decider-2b$^\dagger$ & .59 & .50 & $\mathbf{.65}$ & $\mathbf{.61}$ & $+.06$ {\scriptsize[.02,.09]} & .59 \\
Kev-0.8B$^\dagger$ & .59 & .50 & $\mathbf{.64}$ & $\mathbf{.59}$ & $+.04$ {\scriptsize[$-$.01,.09]} & .59 \\
Laya$^\dagger$ & .52 & .50 & $\mathbf{.62}$ & $\mathbf{.60}$ & -- & -- \\
\midrule
Kev-9B$^\dagger$ & .74 & .68 & $\mathbf{.83}$ & $\mathbf{.83}$ & $+.09$ {\scriptsize[.05,.12]} & .59 \\
decider-12b$^\dagger$ & .69 & .61 & $\mathbf{.80}$ & $\mathbf{.77}$ & $+.11$ {\scriptsize[.08,.14]} & .59 \\
\bottomrule
\end{tabular}
\end{table}

\paragraph{Detection baselines.}
Table~\ref{tab:stop} compares the model's original decision with thresholding its own score,
asking a second yes/no question, contrastive decoding using a screen-removed pass,
and \sysname{}. The gate has the highest balanced accuracy in each reported phone-model row.
For decider-4b, the percentage of failed steps ending in \texttt{complete} decreases from
79\% with the original choice to 38\% with \sysname{}; for Kev-4B it decreases from
86\% to 38\%. These are offline decision-level measurements, not task-success rates.

\begin{table}[h]
\caption{Gating the phone stop decision on held-out apps (balanced accuracy), as in Table~\ref{tab:offline3}, for all seven
models and one more detector. \emph{Model's choice}: no gate. \emph{Own score} and \emph{Second q.}: as in
Table~\ref{tab:offline3}. \emph{Contrastive}: contrastive decoding against a pass without the screen~\cite{shi2024cad}.
Bottom: failed steps that still end in \texttt{complete}, over both failed variants (\textsc{Logged} and
\textsc{Logged+claim}, 1{,}280 states); Table~\ref{tab:effects3} counts \textsc{Logged+claim} only, hence its higher rates.}
\label{tab:stop}
\centering\small
\setlength{\tabcolsep}{4pt}
\begin{tabular}{@{}lccccc@{}}
\toprule
Model & Model's choice & Own score & Second q. & Contrastive & \sysname{} \\
\midrule
decider-4b & .59 & .59 & .60 & .60 & $\mathbf{.71}$ \\
Kev-4B & .59 & .60 & .61 & .61 & $\mathbf{.72}$ \\
decider-2b & .57 & .55 & .59 & .56 & $\mathbf{.64}$ \\
Kev-0.8B & .53 & .56 & .58 & .56 & $\mathbf{.61}$ \\
Laya & .50 & .53 & .55 & .49 & $\mathbf{.59}$ \\
Kev-9B & .66 & .66 & .68 & .65 & $\mathbf{.75}$ \\
decider-12b & .60 & .63 & .62 & .61 & $\mathbf{.73}$ \\
\midrule
\multicolumn{6}{@{}l}{\emph{Wrong stops, decider-4b}: model's choice 79\% $\to$ own score 72\% $\to$ \sysname{} 38\%} \\
\multicolumn{6}{@{}l}{\emph{Wrong stops, Kev-4B}: model's choice 86\% $\to$ own score 58\% $\to$ \sysname{} 38\%} \\
\bottomrule
\end{tabular}
\end{table}

\paragraph{Backbone comparison.}
Table~\ref{tab:backbone} compares a Qwen3.5-4B backbone before and after decision-specific
training on the same 1{,}000 phone states. Both backbone variants are more sensitive to
an explicit claim than to the observation. Decision tuning changes the size of individual
effects but does not consistently correct this imbalance. The comparison describes a
shared pattern in the evaluated checkpoints; it does not establish which training stage
originally caused the preference.

\begin{table}[h]
\caption{The bias is already present in the base model: one Qwen3.5-4B backbone before (top) and after (bottom) decision tuning, 1{,}000 phone
states. Left: paired effect on $\log p(\texttt{complete})$. Right: AUROC for detecting the failure from the output and from
a probe.}
\label{tab:backbone}
\centering\small
\setlength{\tabcolsep}{4pt}
\begin{tabular}{@{}lccc@{\hskip 10pt}cc@{}}
\toprule
Model & Observation & Logged step & Claim & Output & Probe \\
\midrule
Qwen3.5-4B-Base & $+.02$ & $+.15$ & $+.51$ & .59 & .65 \\
Qwen3.5-4B (chat) & $+.05$ & $+.28$ & $+.42$ & .61 & .72 \\
\midrule
Kev-4B (LoRA) & $+.09$ & $+.21$ & $+.32$ & .66 & .77 \\
decider-4b (full fine-tuning) & $+.01$ & $+1.12$ & $+.78$ & .62 & .75 \\
\bottomrule
\end{tabular}
\end{table}

\begin{figure}[h]
\centering
\includegraphics[width=0.55\textwidth]{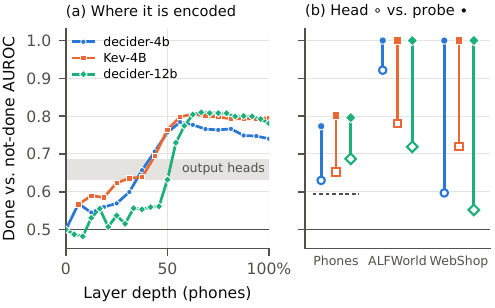}
\caption{Layerwise decoding of action outcomes. (a) Phone probe AUROC across relative depth, compared with the decision model's output (grey band). (b) Model-output AUROC (hollow points) and probe AUROC (filled points) across domains; dashed grey indicates the input-text baseline. In ALFWorld and WebShop, standardized failure messages also allow a text-only classifier to reach AUROC 1.00.}
\label{fig:layers}
\end{figure}

\begin{figure}[h]
\centering
\includegraphics[width=0.45\textwidth]{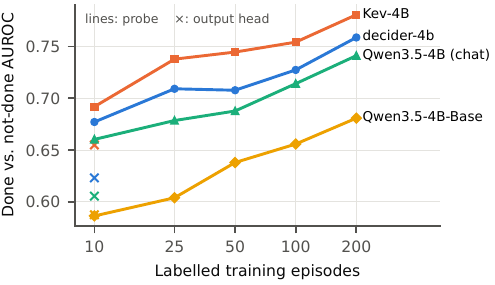}
\caption{How many labels the phone probe needs (held-out apps). $\times$: the output's AUROC. Ten labelled episodes already
beat the output for the decision-tuned models.}
\label{fig:lc}
\end{figure}

\paragraph{Layer depth and label efficiency.}
Figure~\ref{fig:layers} reports AUROC as a function of representation depth.
On phones, the outcome becomes more readily linearly decodable around the middle of the
network. The very high ALFWorld and WebShop probe results should be interpreted alongside
their fixed textual failure traces: a text-only classifier can also separate those outcomes.
Figure~\ref{fig:lc} shows that a phone probe fitted on ten labelled episodes already
outperforms the decision model's output on the reported learning-curve setting. These
results motivate a lightweight readout, without implying that it will transfer to new
state distributions without recalibration.

\section{Removing the Record}
\label{app:blind}

Activation patching in \S\ref{sec:why} identifies where controlled input changes begin to
affect the decision-position representation. We additionally test whether the record's
late influence can be prevented by removing its tokens from subsequent computation.
For decider-4b, we preserve activations through layer $\ell-1$ and re-run layers
$\ell,\ldots,32$ after removing record tokens, while retaining the positions of the
remaining tokens. Removed tokens are phone step summaries, ALFWorld action and thought
entries, or WebShop action entries. We delete rather than mask these tokens because 24
of the 32 layers use recurrent linear attention, for which the standard attention mask
would not implement the intended intervention. As a control, re-running with no token
removal exactly reproduces the normal output.

Table~\ref{tab:blind} reports the resulting rate of premature commitment on failed
steps in the evaluated ALFWorld and WebShop subsets. Removing record tokens beginning
at layer 12 reduces premature advancement in ALFWorld, but removal beginning at layer
20 has no measurable effect in either domain. One plausible explanation is that the
record's influence has already propagated into retained representations by that depth.
Because this intervention changes a partially computed sequence rather than cleanly
isolating a single source of information, it does not by itself prove literal erasure
or overwriting of observational evidence. The direct mid-layer detector is more effective
on these subsets.

\begin{table}[h]
\caption{Failed steps that still get a terminal decision when the record is removed from layer $\ell$ onward (decider-4b),
compared with \sysname{}. Failed steps are both failed variants (\textsc{Logged} and \textsc{Logged+claim}: ALFWorld 590,
WebShop 600 states); Table~\ref{tab:effects3} counts \textsc{Logged+claim} only.}
\label{tab:blind}
\centering\small
\begin{tabular}{@{}lcc@{}}
\toprule
& ALFWorld & WebShop \\
\midrule
Normal pass & 59\% & 99.7\% \\
Record removed from layer 12 & 26\% & 93\% \\
Record removed from layer 16 & 50\% & 99.5\% \\
Record removed from layer 20 & 59\% & 99.7\% \\
\sysname{} & 0.5\% & 0\% \\
\bottomrule
\end{tabular}
\end{table}

\paragraph{Removing the record from the input.} A simpler alternative is to delete the agent's record before the decision
model sees it. We run decider-4b with all record tokens removed from the input (the same tokens as above, removed from layer
1 on) and treat a terminal option as vetoed when the record-free model does not choose it. Table~\ref{tab:norecord} compares
this with \sysname{} on the paired states (phones: the 1{,}000-state subset). Without the record, the phone model loses track
of what it has already done: it rarely chooses \texttt{complete} at all, vetoing 93\% of correct completions, and its
probability of completion no longer separates finished from failed steps (AUROC 0.53). In ALFWorld and WebShop, where the
failure is visible in the observation, removing the record helps, but more failed steps still get through and more good
steps are vetoed than with \sysname{}. The record carries progress information the model needs; the problem is that it also
overrides the observation.

\begin{table}[h]
\caption{Removing the record from the input vs.\ \sysname{} (decider-4b, paired states). \emph{Failed pass}: failed steps
that still end in a terminal decision. \emph{Good vetoed}: steps that worked but whose terminal decision is blocked. Failed and good steps include both variants
with and without the claim (phones: the 1{,}000-state subset, 400 failed and 400 good states; ALFWorld 590 and 590; WebShop
600 and 600).}
\label{tab:norecord}
\centering\small
\setlength{\tabcolsep}{5pt}
\begin{tabular}{@{}lcc@{\hskip 10pt}cc@{\hskip 10pt}cc@{}}
\toprule
& \multicolumn{2}{c}{Phones} & \multicolumn{2}{c}{ALFWorld} & \multicolumn{2}{c}{WebShop} \\
\cmidrule(lr){2-3}\cmidrule(lr){4-5}\cmidrule(l){6-7}
& Failed pass & Good vetoed & Failed pass & Good vetoed & Failed pass & Good vetoed \\
\midrule
No gate & 78\% & 0\% & 59\% & 0\% & 99.7\% & 0\% \\
Record removed from the input & 2\% & 93\% & 9.5\% & 13\% & 6.0\% & 7.5\% \\
\sysname{} & 25\% & 27\% & 0.5\% & 0\% & 0\% & 0.5\% \\
\bottomrule
\end{tabular}
\end{table}

\section{Activation Patching in decider-12b}
\label{app:patch12b}
We repeat the patching of Section~\ref{sec:why} on decider-12b (Gemma-4, 48 layers) with the same procedure and the same
filter ($\log p$ differs by more than 0.2), which leaves 48--94 pairs per curve. Table~\ref{tab:patch12b} shows the same
order as in decider-4b: the observation reaches the decision position first and the record a few layers later. The gap is
largest in WebShop and smallest on phones.

\begin{table}[h]
\centering
\caption{Activation patching in decider-12b (48 layers). Each cell is the share (\%) of an input's effect on the decision
that is restored by patching the decision-position state at that layer. Obs = the screen or observation; logged = the
logged step; claim = the false success claim. Before layer 26 every share is at most 2\%.}
\label{tab:patch12b}
\small
\begin{tabular}{@{}lccc@{}}
\toprule
Layer & Phones & ALFWorld & WebShop \\
 & obs / logged / claim & obs / logged / claim & obs / logged / claim \\
\midrule
28 & 23 / 6 / 5 & 7 / 4 / 2 & 5 / 1 / 3 \\
32 & 66 / 42 / 38 & 50 / 31 / 22 & 46 / 10 / 22 \\
36 & 82 / 68 / 67 & 70 / 50 / 40 & 69 / 32 / 44 \\
42 & 95 / 85 / 85 & 87 / 74 / 67 & 88 / 63 / 73 \\
\bottomrule
\end{tabular}
\end{table}

\section{Serving Latency}
\label{app:serving}

The decision-model forward pass already computes the hidden state used by
\sysname{}. The gate extracts the answer-position representation at one
intermediate layer and evaluates a linear readout, requiring approximately
$2d$ floating-point operations for a dot product in $d$ dimensions
(about $5\times10^3$ for $d=2560$), versus approximately
$2.4\times10^{13}$ operations for a 3,000-token forward pass of the 4B model.
No second model inference is required.

We integrate the readout into decider's serving engine using
\texttt{torch.compile}: CUDA graphs for inputs up to 2{,}048 tokens and a
fused eager path for longer inputs. A hook at the selected layer exposes
the probe score as an additional output column, without an additional
kernel launch or synchronization in this implementation. Table~\ref{tab:overhead}
reports median latency (p50) on one RTX A6000 GPU using bf16,
CUDA-event timing, and 30 iterations after warm-up. Inputs include
actual T3A request states and shorter 280-token states. Added latency
ranges from 0.04\% to 2.01\% in the reported decider conditions.
By comparison, the second-question baseline requires 93--98\% additional
latency, and contrastive decoding requires 56--79\%.
In Kev's serving path, median latency rises from 1207.8 to 1211.2\,ms
(0.29\%). These are measurements of the specific implementations and
hardware configurations tested, not a hardware-independent bound.

\begin{table}[h]
\caption{Decision latency (p50, ms) of decider-4b in its production engine, with and without \sysname{} (30 iterations after
warm-up). The p99 latency of the base engine is within 0.3--0.7\% of p50 for T3A states and 0.7--4.5\% for short states, so
overheads below about 1\% are within measurement variability.}
\label{tab:overhead}
\centering\small
\setlength{\tabcolsep}{6pt}
\begin{tabular}{@{}llrrr@{}}
\toprule
States & Batch & Engine & +\sysname{} & Overhead \\
\midrule
T3A ($\sim$2.7--3.4k tokens) & 1 & 363.4 & 368.3 & $+1.35\%$ \\
 & 8 & 3870.2 & 3873.8 & $+0.09\%$ \\
 & 32 & 19503.6 & 19512.3 & $+0.04\%$ \\
Short ($\sim$280 tokens) & 1 & 39.7 & 40.5 & $+2.01\%$ \\
 & 8 & 214.9 & 217.8 & $+1.32\%$ \\
 & 32 & 829.3 & 831.4 & $+0.26\%$ \\
\bottomrule
\end{tabular}
\end{table}

\section{Complete Phone-Simulator Results}
\label{app:loops}

The phone simulator contains 24 tasks, each evaluated with 30 matched random
seeds (720 trials per condition). A task requires the agent to navigate to
the appropriate app and perform its state-changing action; the record generator
may incorrectly report that a failed action succeeded. The same task--seed
pairs are used to compare conditions. Besides the main-text baselines, we
consider contrastive decoding and a separate 9B LLM judge
(Qwen3.5-9B) queried whenever the decision model proposes stopping.

Table~\ref{tab:e2e} reports task success, paired success differences with
95\% bootstrap intervals over tasks, premature-stop frequency, and inference
calls. The balanced \sysname{} threshold increases success from 77.2\% to
82.4\% ($+5.1$ percentage points, 95\% CI $[1.8,8.9]$) while reducing
false stops from 13.9\% to 8.6\%. The second question performs similarly
in success (83.2\%) but uses more calls per successful episode (6.10 versus
4.86 for \sysname{}). The separate 9B judge achieves 81.5\% success,
requiring 26\% more calls and 29\% more time per successful task than the
ungated baseline in this implementation. Random vetoing offers little
improvement. The simulator's simplified task dynamics and template records
limit how directly these results transfer to real phone interactions.

\begin{table}[h]
\caption{Phone simulator, all conditions (24 tasks $\times$ 30 seeds, decider-4b, \sysname{} with the balanced threshold).
\emph{Gain}: paired change in success vs.\ no gate, with 95\% CI over tasks. \emph{False stops}: episodes that stop before
the goal is reached.}
\label{tab:e2e}
\centering\small
\setlength{\tabcolsep}{5pt}
\begin{tabular}{@{}lccccc@{}}
\toprule
Condition & Success & Gain & False stops & Calls / episode & Calls / success \\
\midrule
No gate & 77.2 & -- & 13.9 & 3.80 & 4.92 \\
Random veto, second question's rate & 77.5 & $+0.3$ {\scriptsize[$-$0.3, 0.8]} & 13.3 & 3.99 & 5.14 \\
Random veto, \sysname{}'s rate & 77.6 & $+0.4$ {\scriptsize[0.0, 0.8]} & 13.3 & 3.98 & 5.12 \\
Contrastive decoding & 77.1 & $-0.1$ {\scriptsize[$-$9.7, 7.9]} & 8.2 & 8.93 & 11.58 \\
Second question & \textbf{83.2} & $+6.0$ {\scriptsize[1.7, 11.2]} & 7.8 & 5.07 & 6.10 \\
LLM judge (9B) & 81.5 & $+4.3$ {\scriptsize[1.7, 7.6]} & 9.3 & 5.04 & 6.18 \\
\sysname{} & 82.4 & $+5.1$ {\scriptsize[1.8, 8.9]} & 8.6 & 4.00 & \textbf{4.86} \\
\bottomrule
\end{tabular}
\end{table}

\section{Fine-Tuning Details}
\label{app:ft}

We fine-tune \texttt{Mapika/decider-4b}, a Qwen3.5-4B model with 32 layers
(24 recurrent linear-attention layers and eight full-attention layers) and
hidden dimension 2560. The base weights remain frozen. We train rank-16
LoRA adapters on all linear layers (scale 32, dropout 0.05), with 32.5 million
trainable parameters, or approximately 0.8\% of the model. We merge the
adapter before evaluation.

Each training example is one of the five paired states and uses its domain's
ordinary decision question. We assign $y=1$ to \textsc{Done} and
\textsc{Done+claim} and $y=0$ to the other three variants. Let $p_T$ denote
the probability of the commitment option, and let $q$ and $q_{\mathrm{base}}$
be the current and original distributions over non-commitment options.
Training combines a binary cross-entropy objective,
\[
 \mathcal{L}_{\mathrm{commit}}=-y\log p_T-(1-y)\log(1-p_T),
\]
with a distribution-preservation term,
\[
 \mathcal{L}_{\mathrm{retain}}=
 \mathrm{KL}\!\left(q_{\mathrm{base}}\,\|\,q\right).
\]
The second term discourages changes to preferences among non-commitment
options while the first trains the commitment score to reflect the outcome.

We use AdamW with learning rate $10^{-4}$, $\beta_1=0.9$,
$\beta_2=0.95$, zero weight decay, 20 warm-up steps followed by cosine
decay, batch size eight, one epoch, gradient clipping at 1.0, bf16,
and random seed 0. Training uses one RTX A6000 GPU.
In the phone-only condition, 2{,}755 states from 205 apps are used for
training, and 445 states from other apps are held out. In the three-domain
condition, training draws 1{,}000 states per domain from phone training
apps, ALFWorld \texttt{train}-split games (excluding the 117 solved test
games), and WebShop goals numbered 3000 and above (excluding test goals).
A third condition uses all 2{,}755 phone training states with the same
1{,}000 ALFWorld and 1{,}000 WebShop states, with identical settings, and is
evaluated on the same held-out apps, games and goals.
To check for broad changes in performance, we also evaluate the first
400 examples from decider's held-out HWU64 and DBpedia-L2 tasks.

\section{Using the Probe as a Best-of-$N$ Verifier}
\label{app:bon}

To test whether the probe is useful beyond binary vetoing, we use it to rank
complete WebShop attempts. For each of 50 goals, the ungated live agent
samples 16 trajectories, generating thoughts at temperature 0.8. For each
candidate attempt that reaches a purchase, we evaluate the decision model's
own answer to ``is the purchased product correct?'' and the probe score,
which was trained on other goals to predict whether required options were
selected. We compare their ability to select a successful attempt, along
with random selection and an oracle that knows the evaluation outcome.

For each value of $N$, we average over 300 randomly selected subsets of
$N$ attempts per goal. Table~\ref{tab:bon} shows that the probe selects a
successful attempt more often than the model's own judgment, with the
largest observed gap at $N=16$: 40.0\% versus 28.0\%, a 12.0-point
advantage (95\% CI $[2.0,22.0]$). The oracle reaches 44.0\% at the
same sampling budget. The probe's training target is option-selection
completion, not general product correctness; these results demonstrate
useful ranking within the evaluated goal distribution, rather than a
general-purpose guarantee of purchase correctness.

\begin{table}[h]
\caption{Best-of-$N$ success (\%) on 50 WebShop goals, using 16 sampled agent trajectories per goal. Each non-oracle strategy selects among candidate attempts using the indicated ranking criterion; results average over 300 random subsets per goal. Strategies choose only among attempts that end in a purchase, and a goal counts as failed when none of its $N$ attempts buys; this is why even random selection improves with $N$. Confidence intervals resample goals.}
\label{tab:bon}
\centering\small
\begin{tabular}{@{}lccccc@{}}
\toprule
$N$ & 1 & 2 & 4 & 8 & 16 \\
\midrule
Oracle & 9.8 & 16.9 & 25.9 & 35.8 & 44.0 \\
Random attempt & 9.8 & 14.1 & 17.2 & 19.8 & 21.2 \\
Decision model's own judgement & 9.8 & 15.2 & 20.4 & 24.9 & 28.0 \\
Probe on its hidden state & 9.8 & 16.5 & 24.3 & 32.8 & 40.0 \\
\bottomrule
\end{tabular}
\end{table}

\end{document}